\documentclass{article}
\usepackage{todonotes}
\usepackage{arxiv}
\usepackage{float}
\usepackage[utf8]{inputenc}
\usepackage[T1]{fontenc}

\usepackage{times}
\usepackage{helvet}
\usepackage{courier}
\usepackage{placeins}
\usepackage{hyperref}
\usepackage{url}
\usepackage{booktabs}
\usepackage{amsfonts}
\usepackage{amsmath}
\usepackage{amssymb}
\usepackage{nicefrac}
\usepackage{microtype}
\usepackage{graphicx}
\usepackage{natbib}
\usepackage{doi}
\usepackage{multirow}
\usepackage{array}
\usepackage{caption}
\usepackage{subcaption}

\usepackage{amsthm}

\theoremstyle{definition}
\newtheorem{definition}{Definition}

\usepackage{algorithm}
\usepackage{algorithmic}

\title{A Benchmark for Procedural Memory Retrieval in Language Agents}

\author{
  Ishant K \\
  Qpi AI \\
  \texttt{ishant@qpiai.tech} \\
  \And
  Aswanth Krishnan \\
  Qpi AI \\
  \texttt{ashwanth.krishnan@qpiai.tech} \\
}

\renewcommand{\shorttitle}{Procedural Memory Retrieval Benchmark}

\hypersetup{
pdftitle={Benchmarking Procedural Memory Retrieval: The Generalization Cliff in Cross-Context Understanding},
pdfsubject={cs.AI, cs.LG},
pdfauthor={Ishant K, Aswanth Krishnan},
pdfkeywords={Procedural Memory, Retrieval Benchmarking, Cross-Context Generalization, AI Agents, ALFWorld},
}

\begin{document}
\maketitle

\begin{abstract}
Current AI agents excel in familiar settings, but fail sharply when faced with novel tasks with unseen vocabularies—a core limitation of procedural memory systems. We present the first benchmark that isolates procedural memory retrieval from task execution, evaluating whether agents can recognize functionally equivalent procedures that span different object instantiations. Using ALFWorld, we construct dual corpora of expert and LLM-generated trajectories and evaluate six retrieval methods using systematically stratified queries. Our results expose a clear generalization cliff: embedding-based methods perform strongly on familiar contexts, yet degrade considerably on novel ones, while LLM-generated procedural abstractions demonstrate reliable cross-context transfer. Controlled ablations show that although embeddings capture some lexical-level abstraction, they fundamentally treat procedures as unordered bags of words, discarding temporal structure necessary for cross-context transfer. Corpus scale delivers far larger gains than representation enrichment, revealing an architectural ceiling in current encoders. Our benchmark offers the first diagnostic framework separating genuine procedural understanding from surface-level memorization and gives tools for developing retrieval systems capable of dependable generalization. Resources available at \href{https://github.com/qpiai/Proced_mem_bench}{our GitHub repository}.
\end{abstract}

\section{Introduction}

AI agents increasingly depend on retrieving relevant procedural knowledge from past experience to solve new tasks. For example, an agent who has learned to ``clean an apple with water and place it in a cabinet'' should recognize that ``cleaning a salt shaker with water and placing it in a drawer'' follows the same underlying procedure. This capability of retrieving procedurally similar trajectories that span different object contexts is necessary for generalization, transfer, and sample-efficient learning. However, current evaluation frameworks cannot reliably distinguish systems that truly understand procedural structure from those that simply memorize surface-level lexical patterns.

Existing work on memory-augmented agents, trajectory retrieval, and retrieval-augmented generation \citep{shinn2023reflexion, lewis2020rag} demonstrates that incorporating prior experience can improve downstream task performance. However, the available benchmarks mainly evaluate the end-to-end success, conflating retrieval with planning, grounding, and execution. An agent may appear to retrieve well by exploiting familiar vocabulary, or fail due to poor action generation despite retrieving appropriate demonstrations. What is missing is an evaluation framework that isolates \emph{retrieval quality}: the ability to identify relevant procedures based solely on structural similarity.

We address this gap by introducing a benchmark that systematically evaluates procedural memory retrieval independently of execution. Our benchmark is built on ALFWorld \citep{shridhar2021alfworld}, a text-based environment that offers various household tasks with controllable complexity and clearly defined action sequences. This domain lets us focus explicitly on procedural understanding rather than visual grounding or motor control.

To ensure robustness and generality, we evaluate retrieval systems in two complementary corpora: (1) an authentic dataset of expert demonstrations generated by ALFWorld's HandCodedTWAgent, capturing realistic exploration behavior, and (2) a large LLM-generated corpus from AgentInstruct \citep{agentinstruct2024dataset}, offering scale and diverse procedural variations. This dual-corpus approach supports cross-validation of findings spanning data sources and generation methods.

\begin{figure}[t]
\centering
\includegraphics[width=\linewidth]{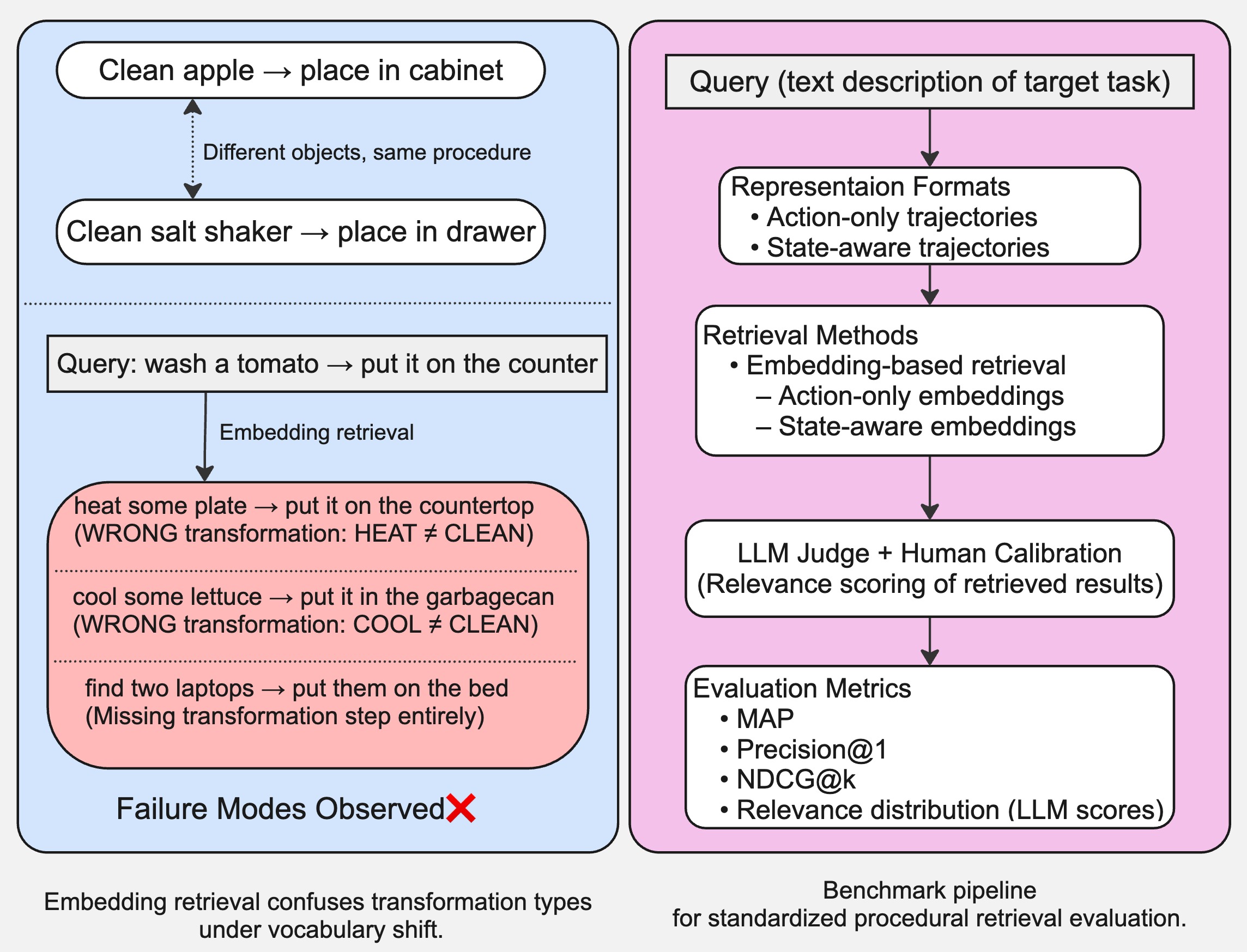}
\caption{
Overview of the procedural retrieval problem and benchmark pipeline.
\textbf{Left:} Embedding-based retrieval confuses transformation types under vocabulary shift
(e.g., CLEAN vs.\ HEAT vs.\ COOL), producing structurally incorrect matches.
\textbf{Right:} Our benchmark pipeline evaluates retrieval quality via action-only and state-aware
representations, LLM-judged relevance scoring, and standardized IR metrics.
}
\label{fig:overview}
\end{figure}

Figure~\ref{fig:overview} illustrates the core challenge motivates our benchmark and the high-level evaluation pipeline used throughout this work.

Our evaluation framework combines two complementary regimes. First, an exploratory dual-condition analysis contrasts performance on familiar queries versus novel ones to uncover distribution shifts. Second, we develop a coverage-balanced benchmark that systematically controls for corpus representation by selecting queries with guaranteed procedural coverage and stratifying them into tiers of task complexity. This prevents confounds such as spurious correlations between task difficulty and corpus frequency.

Throughout both regimes, our experiments uncover a striking distribution shift failure. Retrieval methods that perform strongly on familiar contexts experience marked degradation on novel ones, reversing the ranking of embedding methods and exposing an architectural trade-off between memorization and abstraction. In contrast, LLM-generated procedural summaries which explicitly remove object-specific information display far more stable behavior, suggesting that abstraction at the representation level plays a key role in cross-context transfer.

Finally, we validate our findings through five complementary analyses, including lexical overlap, corpus ablation, semantic task-space clustering, controlled format comparison, and manual human annotation. These studies establish that current sentence-transformer encoders behave largely as ``bag-of-words'' models on procedural data, limiting their ability to capture temporal structure and causal dependencies.

\subsection{Contributions}

Our contributions are:

\begin{enumerate}
    \item A benchmark that isolates procedural memory retrieval from end-to-end execution, including dual corpora, a coverage-balanced query bank, and a validated evaluation protocol.
    \item Discovery and characterization of the \emph{generalization cliff}, demonstrating that embedding-based retrieval systems strongly overfit to familiar object vocabularies and struggle to generalize procedural patterns.
    \item A comprehensive diagnostic suite establishing that current embedding architectures prioritize memorization over abstraction, with corpus scale dominating representation quality.
    \item Actionable insights for retrieval system design, motivating structure-aware representations and two-stage retrieval architectures that separate procedural extraction from similarity computation.
    \item Open-source release of datasets, methodology, and evaluation framework to support future research on dependable procedural memory systems.
\end{enumerate}

\section{Related Work}
\label{sec:related}

\subsection{Embodied Benchmarks and Procedural Memory Systems}

Embodied AI benchmarks such as ALFWorld \citep{shridhar2021alfworld} evaluate agents through end-to-end task completion, making it difficult to distinguish whether failures arise from retrieval, planning, or execution. Recent procedural memory systems attempt to make memory more explicit. Memp \citep{fang2025mempexploringagentprocedural} builds a lifelong procedural memory from agent trajectories, storing both step-level instructions and higher-level abstractions, and confirms improved task efficiency and transfer spanning models. LEGOMem \citep{han2025legomemmodularproceduralmemory} introduces a modular multi-agent memory architecture where an orchestrator manages full-task memories while specialized agents retrieve subtask memories, improving workflow automation. These systems underscore the value of procedural memory but still assess performance through downstream task success rather than retrieval quality.

Broader memory-focused frameworks stress the structural limitations of current agents. Wheeler and Jeunen \citep{wheeler2025procedural} argue that relying solely on procedural memory constrains adaptability in environments with shifting rules, advocating stronger integration of semantic and associative memory. MIRIX \citep{wang2025mirixmultiagentmemoryllmbased} operationalizes this through six coordinated memory types (Core, Episodic, Semantic, Procedural, Resource, Knowledge Vault) managed by a meta-memory controller, supporting richer multimodal memory usage. While these approaches expand the role of memory, they do not directly evaluate whether agents recognize procedurally similar trajectories spanning different object contexts.

\subsection{Retrieval Architectures and Memory Access Methods}

Most agent retrieval mechanisms rely on semantic encoders or lexical baselines. Sentence-BERT \citep{reimers2019sentence} is still a widely used embedding model for semantic search, while BM25 \citep{robertson2009probabilistic} delivers strong lexical retrieval despite its simplicity. More recent work explores retrieval architectures that better integrate temporal or multimodal information. TRACE \citep{chen2025tracegroundingtimeseries} grounds time-series embeddings in aligned textual context, supporting cross-modal retrieval and state-of-the-art results on forecasting and classification tasks. Although not designed for procedural trajectories, it underscores the importance of aligning temporal patterns with semantic signals.

Several agent systems incorporate retrieval modules more directly. Hong and He \citep{hong2025memory} propose LLM-trained cross-attention networks to improve memory retrieval for generative agents, confirming more consistent behavior in simulated environments. PAL-UI \citep{palui2025} lets GUI agents retrieve past screenshots through an active look-back mechanism for long-horizon planning. These methods focus on improving retrieval effectiveness inside agent pipelines but do not measure retrieval quality independently from downstream planning and execution.

\subsection{Trajectory Datasets and Agent Training}

Large-scale trajectory datasets support training and evaluation of language agents but do not isolate retrieval performance. AgentInstruct \citep{agentinstruct2024dataset} offers GPT-4–generated trajectories filtered for successful execution, PAL-UI \citep{palui2025} collects 8.6K step-level instructions for GUI navigation, and TrajAgent \citep{du2025trajagentllmagentframeworktrajectory} introduces trajectory representations optimized for few-shot generalization. Voyager \citep{voyager2023} builds an ever-growing skill library that functions as procedural memory for exploration. Although these datasets improve agent learning, they assume that retrieval works well and never evaluate retrieval quality directly.

Survey papers \citep{zhang2024memorysurvey} stress that memory is central to supporting self-improving agents, yet existing evaluations assess memory implicitly through final task success. Consequently, it's unclear whether improved performance comes from better retrieval, better reasoning, or better action generation.

\subsection{Cross-Context Generalization and Scaling Behavior}

Generalization spanning contexts is a central challenge. Prior work explores improvements in trajectory prediction \citep{xu2025trajectorypredictionmeetslarge}, long-horizon context compression \citep{kang2025aconoptimizingcontextcompression}, and multimodal agent memory organization \citep{coala2024}. Yet these approaches focus on policy or representation generalization rather than retrieval generalization—whether a system can identify procedurally similar trajectories when vocabularies shift.

Scaling laws \citep{kaplan2020scaling} point to logarithmic performance improvements with data size. Recent agent memory work aligns with this observation: larger procedural corpora tend to yield more dependable behavior, and context engineering frameworks \citep{langchain2025} stress that agents depend on the quality of retrieved context. But no existing study evaluates how retrieval quality itself scales with corpus size or changes under distribution shifts.

\subsection{Missing Benchmark: Retrieval Quality Under Distribution Shift}

Despite considerable progress, no existing work isolates procedural memory retrieval from end-to-end task performance or evaluates retrieval quality under cross-context distribution shifts. Prior systems focus on constructing, organizing, or leveraging memory, but none test whether a system can recognize that two trajectories with different object vocabularies instantiate the same underlying procedure. Our work fills this gap by offering the first benchmark dedicated to procedural retrieval quality, evaluating whether agents retrieve functionally equivalent past experiences when surface features differ.

\section{Methodology}
\label{sec:methodology}

\subsection{Formal Problem Definition}

We formally define the procedural memory retrieval problem as follows:

\begin{definition}[Procedural Trajectory]
A procedural trajectory $t \in T$ is a sequence of state-action pairs $t = \langle (s_1, a_1), \ldots, (s_n, a_n) \rangle$ where $s_i \in S$ represents environmental state and $a_i \in A$ represents the action taken.
\end{definition}

\begin{definition}[Procedural Similarity]
Given trajectories $t_i, t_j \in T$, procedural similarity $\text{sim}_\text{proc}: T \times T \rightarrow [0,1]$ measures functional equivalence independent of object vocabularies:
\begin{equation}
\text{sim}_\text{proc}(t_i, t_j) = \alpha \cdot \phi_\text{struct}(t_i, t_j) + (1-\alpha) \cdot \phi_\text{sem}(t_i, t_j)
\end{equation}
where $\phi_\text{struct}$ captures structural patterns (action sequences) and $\phi_\text{sem}$ captures semantic intent.
\end{definition}

\begin{definition}[Cross-Context Retrieval Problem]
Given:
\begin{itemize}
\item Corpus $\mathcal{C} = \{t_1, \ldots, t_N\}$ of procedural trajectories
\item Query $q$ representing a novel task with unseen object vocabulary $V_q \cap V_\mathcal{C} = \emptyset$
\item Retrieval function $f: Q \times \mathcal{C} \rightarrow \mathcal{P}(\mathcal{C})$ returning ranked subset
\end{itemize}

\noindent\textbf{Objective:} Maximize expected procedural utility:
\begin{equation}
\mathbb{E}_{q \sim Q_\text{novel}}[\text{MAP}(f(q, \mathcal{C}), \text{rel}_\text{proc}(q))]
\end{equation}
where $\text{rel}_\text{proc}(q)$ denotes procedurally relevant trajectories for query $q$.
\end{definition}

\begin{definition}[Generalization Gap]
For retrieval method $M$, the generalization gap quantifies performance degradation on novel contexts:
\begin{align}
\text{Gap}(M) &= \frac{\text{MAP}_\text{seen}(M) - \text{MAP}_\text{unseen}(M)}{\text{MAP}_\text{seen}(M)} \\
 \text{Robustness}(M) &= 1 - \text{Gap}(M)
\end{align}
\end{definition}

\subsection{Corpus Construction}

We construct dual corpora to support cross-validation of findings and test generalization spanning different data sources and generation methodologies.

\subsubsection{ALFWorld Expert Trajectory Corpus}

We generated 78 authentic expert trajectories using ALFWorld's HandCodedTWAgent executing tasks from the training split. Rather than using synthetic idealized trajectories, we captured realistic agent behavior including exploration patterns (e.g., ``go to shelf 1 $\rightarrow$ go to garbagecan 1 $\rightarrow$ go to dresser 1'' before finding the target object). The corpus comprises 63 successful expert trajectories and 15 exploration-interrupted sequences, offering both gold-standard procedures and realistic partial executions.

We converted trajectories to text-only format, representing each step as a state-action pair: the environmental observation followed by the chosen action. The corpus spans all six ALFWorld task types: \texttt{pick\_and\_place\_simple}, \texttt{look\_at\_obj\_in\_light}, \texttt{pick\_clean\_then\_place\_in\_recep}, \texttt{pick\_heat\_then\_place\_in\_recep}, \texttt{pick\_cool\_then\_place\_in\_recep}, and \texttt{pick\_two\_obj\_and\_place}.

\textbf{Corpus Statistics:} Average trajectory length: 14.2 actions ($\sigma=5.8$), object vocabulary: 42 unique object types, procedural pattern diversity: 6 task types with multiple instantiations per type.

\subsubsection{AgentInstruct State-Action Corpus}

We adopted the 336 ALFWorld trajectories given in the AgentInstruct dataset \cite{agentinstruct2024dataset}, which were filtered from 954 initial GPT-4–generated demonstrations, yielding a 35.2\% retention ratio. We converted AgentInstruct's conversational traces into explicit state–action pairs by pairing each environment observation with the subsequent agent action.

This corpus delivers 4.3$\times$ scale increase over our expert corpus and supports testing whether findings generalize spanning generation methods (rule-based expert versus LLM-generated) and representation richness (action-only versus state-action format).

\textbf{Corpus Statistics:} Average trajectory length: 12.7 actions ($\sigma=4.3$), object vocabulary: 38 unique types, task type distribution consistent with ALFWorld training split.

\subsection{Retrieval Methods Evaluated}

We evaluate six retrieval approaches spanning modern embedding methods and traditional information retrieval baselines.

\subsubsection{Embedding Methods}

All embedding methods use the \texttt{all-MiniLM-L6-v2} sentence transformer (384-dimensional embeddings) \cite{sentence_transformers} for controlled comparison, with trajectories stored in ChromaDB vector database and retrieval via cosine similarity:
\begin{equation}
\text{sim}(q, d) = \frac{q \cdot d}{\|q\|_2 \|d\|_2}
\end{equation}
where $q$ and $d$ are query and document embeddings.

\textbf{Action-Only Embeddings}: Embeds raw action sequences only, testing whether action patterns alone encode procedural knowledge.
\begin{itemize}
    \item \textbf{Format:} ``look, go to toaster 1, go to stoveburner 1, open cabinet 2, take mug 1...''
    \item \textbf{Hypothesis:} Procedural patterns are captured in action sequence structure.
\end{itemize}

\textbf{Enriched Embeddings}: Embeds actions with task metadata and environmental context.
\begin{itemize}
    \item \textbf{Format:}
\begin{verbatim}
Task Type: pick_heat_then_place_in_recep
Task: Place heated plate in fridge
Objects: plate, fridge, microwave
Complexity: 17 actions
Actions: look, go to toaster 1...
\end{verbatim}

    \item \textbf{Hypothesis:} Additional semantic context improves matching.
\end{itemize}

\textbf{Summary Embeddings}: Embeds LLM-generated procedural abstractions that remove specific object references while preserving procedural structure.
\begin{itemize}
\item Generation: GPT-5 \citep{openai2025gpt5} generates high-level summaries abstracting away objects
\item Example: ``Procedure: locate storage location $\rightarrow$ navigate to container $\rightarrow$ retrieve target item $\rightarrow$ locate heating appliance $\rightarrow$ heat item $\rightarrow$ locate cold storage $\rightarrow$ place item''
\item Hypothesis: Object-agnostic abstractions support better generalization
\item Note: Summary embeddings are evaluated only on the 78-trajectory expert corpus, since generating 336 LLM summaries was computationally expensive.
\end{itemize}

\textbf{Combined Embeddings}: Embeds concatenation of context-enriched representation with procedural summary.
\begin{itemize}
\item Format: Concatenation of enriched representation and procedural summary.
\item Hypothesis: Multiple information sources yield robustness
\end{itemize}

\subsubsection{Baseline Methods}

\textbf{BM25:} A classic TF–IDF lexical retrieval baseline. We evaluate two variants:
\begin{itemize}
    \item \textbf{BM25 (State–Action Text):} indexes the full state–action trace.
    \item \textbf{BM25 (Action-Only Text):} indexes the action sequence only, omitting verbose state observations.
\end{itemize}

\textit{Note:} BM25 variants were explored in preliminary experiments, but we omit them from the final benchmark since their evaluation requires using text that directly overlaps with held-out queries, which would introduce information leakage.

\vspace{\baselineskip}

\textbf{Keyword Matching:} Simple lexical overlap using Jaccard similarity:
\begin{equation}
J(A, B) = \frac{|A \cap B|}{|A \cup B|}
\end{equation}
where $A$ and $B$ are token sets from query and trajectory.

\subsection{Evaluation Framework}

Our evaluation uses two complementary approaches: (1) an exploratory two-condition analysis to identify cross-distribution performance patterns, and (2) a coverage-balanced benchmark that validates these patterns while controlling for corpus-induced artifacts.

\subsubsection{Initial Dual-Condition Evaluation}

\textbf{Purpose:} Assess generalization by comparing retrieval performance on in-distribution and out-of-distribution tasks.

\textbf{Query Sets:}

\textit{Seen Queries} ($n=18$): Task instances procedurally aligned with the training corpus and containing only in-vocabulary objects.

\textit{Unseen Queries} ($n=18$): Novel ALFWorld validation tasks with object vocabularies absent from the corpus, testing generalization under distribution shift.

\textbf{Semantic Equivalence Check:} To ensure comparisons are fair, we embedded all 36 queries using the same sentence transformer used for retrieval (\texttt{all-MiniLM-L6-v2}) and compared pairwise similarities. The two query sets display no meaningful semantic difficulty difference, confirming that performance gaps arise from retrieval behavior rather than intrinsic query complexity.

\textbf{Evaluation Method:} For each query, we retrieve $K=5$ trajectories and evaluate relevance using the LLM-as-judge (§\ref{sec:llm_judge}). Standard IR metrics (MAP, NDCG, Precision@$k$, Recall@$k$) measure ranking quality within retrieved pools. This setup supported rapid exploratory comparison spanning six methods and uncovered sharp performance divergence under vocabulary shift.

\textbf{Identified Limitation:} Post-hoc analysis revealed that unseen queries had variable corpus coverage (2-15 relevant trajectories), whereas seen queries consistently had higher coverage. This imbalance, combined with initial complexity-based stratification, produced artifacts such as inverted difficulty tiers driven by vocabulary frequency rather than true procedural difficulty.

\subsubsection{Coverage-Balanced Benchmark Development}

\textbf{Motivation:} To confirm the generalization cliff observed in the exploratory setting, we construct a refined benchmark that removes corpus-coverage imbalance as a confounding factor.

\textbf{Coverage-First Strategy:}

\textit{Phase 1 – Coverage Estimation:} We sample 200 candidate tasks from the ALFWorld validation set and estimate corpus coverage using lightweight keyword matching (noun/verb extraction with a minimum 2-word overlap). Tasks with 8–20 relevant trajectories are retained, ensuring sufficient procedural support (at least 8 examples) without making retrieval trivial (cap at 20).

\textit{Phase 2 – Complexity Stratification:} Coverage-qualified tasks are grouped into three procedural complexity tiers based on object cardinality, presence of state transformations, and multi-step dependencies:

\begin{itemize}
    \item \textbf{EASY} ($n=15$): Single-object placement tasks  
    Example: ``Put a pencil on the desk.''  
    Mean coverage: 14.4 trajectories ($\sigma=3.2$)

    \item \textbf{MEDIUM} ($n=14$): Multi-step operations involving state changes (e.g., heating or cleaning)  
    Example: ``Heat an apple and place it in the fridge.''  
    Mean coverage: 12.5 trajectories ($\sigma=2.8$)

    \item \textbf{HARD} ($n=11$): Multi-object coordination or composite procedures  
    Example: ``Put two bars of soap in the cabinet.''  
    Mean coverage: 14.0 trajectories ($\sigma=2.5$)
\end{itemize}

\textit{Phase 3 – Validation:} A pilot evaluation confirms that the three tiers display the expected spread in retrieval difficulty, validating the stratification.

\textbf{Coverage Guarantee:} All 40 selected queries exceed the 8-trajectory relevance threshold, verified via LLM-based relevance checking. This removes data-scarcity artifacts and ensures that performance differences capture procedural complexity rather than uneven corpus support.

\textbf{Query Bank Composition:} The final benchmark uses all 40 coverage-balanced queries. The 36 queries from the initial dual-condition setting (18 seen, 18 unseen) are retained for cross-method validation and for analyzing the generalization cliff, with partial overlap between the sets.

\subsubsection{State-Aware vs. Action-Only Representation Formats}

To isolate the effect of representation choice, we evaluate the 336 AgentInstruct trajectories using two formats—state-aware and action-only—under the same embedding model (all-MiniLM-L6-v2) and the coverage-balanced 40-query benchmark.

\textbf{State-Aware Embeddings:}
\begin{itemize}
    \item \textbf{Format:} Task description followed by state–action pairs that include environmental observations.
    \item \textbf{Example:} "Task: Clean soap and put in cabinet. Step 1: [State: On countertop] [Action: take soap 1]. Step 2: [State: Holding soap] [Action: go to sinkbasin 1] …"
    \item \textbf{Information Density:} Approximately 10× longer than action-only due to state descriptions.
\end{itemize}

\textbf{Action-Only Embeddings:}
\begin{itemize}
    \item \textbf{Format:} Pure action sequence without environmental context.
    \item \textbf{Example:} "go to countertop 1, take soap 1, go to sinkbasin 1, clean soap 1, go to cabinet 1, put soap 1"
    \item \textbf{Information Density:} Minimal representation focused on procedural steps.
\end{itemize}

This comparison isolates representation effects while holding corpus content, embedding model, query set, and evaluation protocol constant.

\subsubsection{LLM-as-a-Judge Evaluation}
\label{sec:llm_judge}

\textbf{Rationale:} Classical lexical and overlap-based metrics fail on out-of-distribution tasks because they rely on exact object vocabulary matches. To evaluate functional similarity instead of surface overlap, we adopt an LLM-as-a-judge approach.

We treat LLM-based scoring as a scalable proxy for procedural relevance rather than definitive ground truth; human judgments are used to calibrate and validate its behavior.

\textbf{Configuration:}
\begin{itemize}
    \item \textbf{Model:} GPT-5 \citep{openai2025gpt5} with reasoning effort set to ``low'' for efficient, low-variance scoring.
    \item \textbf{Prompting:} A single task-agnostic prompt stressing functional and procedural equivalence.
    \item \textbf{Output Format:} A graded relevance score on a scale of 1 to 10 scale accompanied by brief justification.
\end{itemize}

\begin{figure}[t]
\centering
\includegraphics[width=\columnwidth]{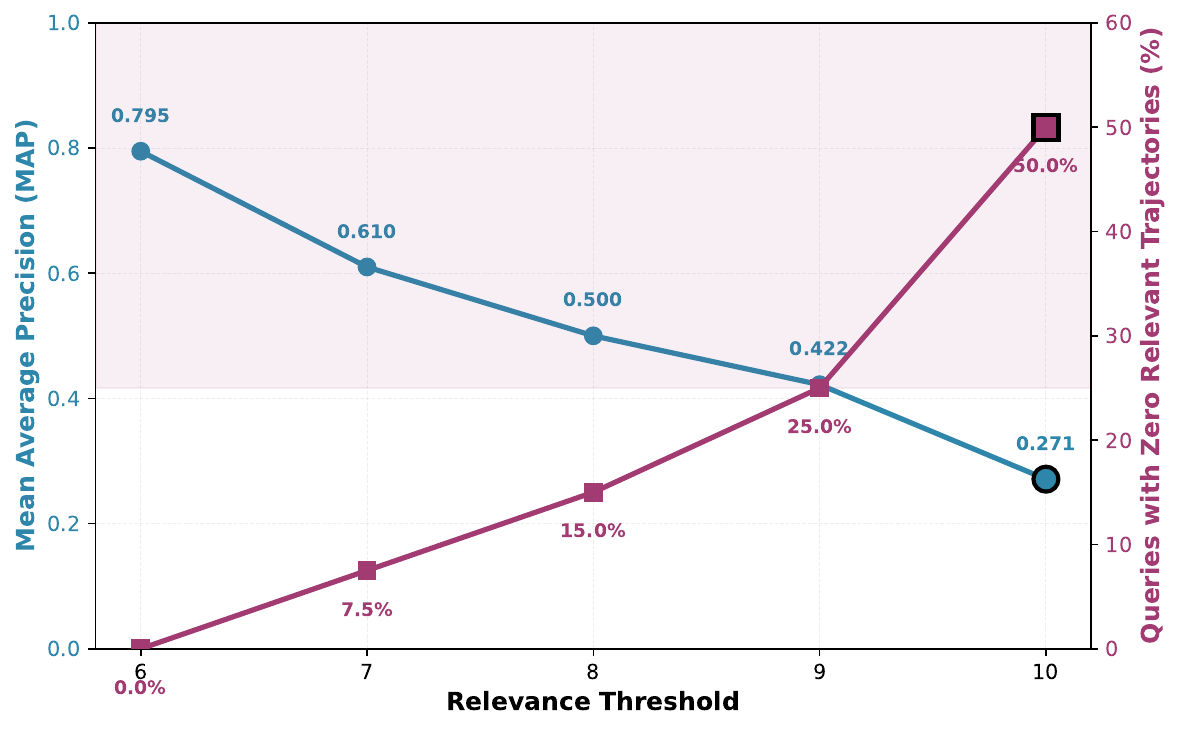}
\caption{Threshold sensitivity analysis documenting MAP performance and percentage of zero-relevant queries for thresholds 6-10. Threshold $\geq 6$ delivers the best balance between retrieval quality and corpus coverage.}
\label{fig:threshold_analysis}
\end{figure}

\textbf{Relevance Threshold Calibration:}  
We conduct threshold sensitivity analysis (6-10) to determine the binary cutoff for relevance. Higher thresholds produce artificial sparsity (Figure-\ref{fig:threshold_analysis}):
\begin{itemize}
    \item $\geq 10$: 50\% queries yield zero relevant trajectories  
    \item $\geq 9$: 25\% zero-relevant  
    \item $\geq 8$: 15\% zero-relevant  
    \item $\geq 7$: 7.5\% zero-relevant  
    \item $\geq 6$: 0\% zero-relevant
\end{itemize}

We adopt threshold $\geq 6$, which avoids induced data scarcity while maintaining a meaningful notion of procedural relevance.

\textbf{Scoring Application:}
\begin{itemize}
    \item \textbf{Binary relevance:} score $\geq 6$ for MAP, Precision@$k$, Recall@$k$, and F1.
    \item \textbf{Graded relevance:} normalized score $(\text{score}/10)$ for NDCG.
    \item \textbf{Evaluation pools:} $K=5$ for dual-condition experiments; $K=10$ for the coverage-balanced benchmark.
\end{itemize}

\subsubsection{Evaluation Metrics}

We report standard information retrieval metrics computed per query and averaged over the query set.

\textbf{Mean Average Precision (MAP):} Primary ranking metric.
\begin{equation}
\text{MAP} = \frac{1}{|Q|} \sum_{q \in Q} \frac{1}{|R_q|} \sum_{k=1}^{N} P@k \cdot \text{rel}_q(k)
\end{equation}
where $R_q$ is the set of relevant trajectories for query $q$, $N$ is the number of retrieved items, and $\text{rel}_q(k)$ is the binary relevance at rank $k$.

\textbf{Precision@k:} Fraction of relevant trajectories in the top-$k$.
\begin{equation}
P@k = \frac{1}{k} \sum_{i=1}^{k} \text{rel}(i)
\end{equation}

\textbf{Recall@k:} Fraction of all relevant trajectories retrieved in the top-$k$.
\begin{equation}
R@k = \frac{1}{|R_q|} \sum_{i=1}^{k} \text{rel}(i)
\end{equation}
where Recall@k is computed over the retrieved pool (not the full corpus), following standard pool-based IR evaluation.

\textbf{F1@k:} Harmonic mean of Precision@k and Recall@k.
\begin{equation}
F1@k = \frac{2 \cdot P@k \cdot R@k}{P@k + R@k}
\end{equation}

\textbf{NDCG@k:} Ranking quality using graded relevance.
\begin{equation}
\text{NDCG}@k = \frac{\text{DCG}@k}{\text{IDCG}@k}, \quad
\text{DCG}@k = \sum_{i=1}^{k} \frac{2^{\text{rel}(i)} - 1}{\log_2(i+1)}
\end{equation}

For complexity-stratified analysis, we additionally report metrics separately for the EASY, MEDIUM, and HARD tiers.

\subsubsection{Coverage-Balanced Query Selection Algorithm}
We implement the coverage-first query selection strategy in two stages: (1) filtering validation tasks by estimated corpus coverage, and (2) stratifying the remaining tasks by procedural complexity.

\begin{algorithm}[htbp]  
\caption{Coverage-Balanced Query Selection}
\label{alg:query_selection}
\begin{algorithmic}[1]
\REQUIRE Validation tasks $V$, Corpus $\mathcal{C}$, Coverage bounds $[m, M]$
\ENSURE Stratified query set $Q = \{Q_\text{easy}, Q_\text{medium}, Q_\text{hard}\}$
\STATE $\text{candidates} \leftarrow \emptyset$
\FOR{each task $t \in V$}
    \STATE $\text{coverage} \leftarrow \textsc{EstimateCoverage}(t,\mathcal{C})$
    \IF{$m \leq \text{coverage} \leq M$}
        \STATE $\text{complexity} \leftarrow \textsc{ClassifyComplexity}(t)$
        \STATE $\text{candidates}[\text{complexity}].\textsc{add}(t)$
    \ENDIF
\ENDFOR
\STATE $Q_\text{easy} \leftarrow \textsc{SampleUniform}(\text{candidates}[\text{EASY}], n_\text{easy})$
\STATE $Q_\text{medium} \leftarrow \textsc{SampleUniform}(\text{candidates}[\text{MEDIUM}], n_\text{medium})$
\STATE $Q_\text{hard} \leftarrow \textsc{SampleUniform}(\text{candidates}[\text{HARD}], n_\text{hard})$
\RETURN $Q_\text{easy} \cup Q_\text{medium} \cup Q_\text{hard}$
\end{algorithmic}
\end{algorithm}

\begin{algorithm}[htbp]  
\caption{Coverage Estimation}
\label{alg:coverage_estimation}
\begin{algorithmic}[1]
\REQUIRE Task $t$, Corpus $\mathcal{C}$
\ENSURE Estimated coverage count
\STATE $\text{keywords} \leftarrow \textsc{ExtractKeywords}(t)$
\STATE $\text{count} \leftarrow 0$
\FOR{each trajectory $\tau \in \mathcal{C}$}
    \IF{$|\text{keywords} \cap \textsc{Tokens}(\tau)| \geq \theta$}
        \STATE $\text{count} \leftarrow \text{count} + 1$
    \ENDIF
\ENDFOR
\RETURN \text{count}
\end{algorithmic}
\end{algorithm}

\textbf{Complexity:} The procedure runs in $O(|V| \cdot |\mathcal{C}| \cdot L)$ time, dominated by keyword matching spanning corpus trajectories during coverage estimation.

\subsection{Validation Studies}

We perform five complementary validation studies to assess the robustness of our findings and to characterize key limitations of current embedding-based retrieval systems.

\subsubsection{Lexical Overlap Analysis}
We examine whether embedding models rely on true semantic abstraction or merely encode lexical similarity. For each query–trajectory pair in the evaluation set, we compute token-level Jaccard overlap and correlate it with LLM-assigned relevance scores. Analyses are conducted separately for seen and unseen settings. A strong correlation would point to keyword-driven behavior; weak correlation would hint at abstraction beyond surface vocabulary.

\subsubsection{Corpus Size Ablation}
To separate corpus-scale effects from representation quality, we compare retrieval performance using two versions of the AgentInstruct corpus: the full 336-trajectory set and a 78-trajectory stratified subsample preserving task-type distribution. Both conditions use identical embeddings (state-aware format, all-MiniLM-L6-v2) and the same query set and evaluation protocol. Differences in MAP and complexity-tier performance isolate the contribution of corpus size.

\subsubsection{State-Aware vs. Action-Only Comparison}
We evaluate how representation richness influences retrieval performance by comparing state-aware embeddings (with environmental context) against action-only embeddings on the same 336 trajectories, using a shared embedding model and the coverage-balanced 40-query benchmark. This isolates representation effects independent of corpus content or evaluation setup. If richer context fundamentally improves procedural understanding, state-aware embeddings should outperform action-only; otherwise, gains should be modest.

\subsubsection{Semantic Task Space Analysis}

We assess whether the seen/unseen split captures genuine semantic differences by embedding 506 ALFWorld task-instruction instances (251 seen, 255 unseen) derived from 327 underlying task configurations, using the same sentence transformer as our retrieval model. We compute pairwise cosine similarities and examine group structure via t-SNE/UMAP visualizations. We quantify separation using within-group vs.\ cross-group similarity distributions and effect-size measures (Cohen's $d$ ) alongside silhouette scores.

\textbf{Interpretation:} The weak separation implies that seen and unseen tasks are semantically similar in embedding space. Consequently, performance differences cannot be attributed solely to intrinsic query difficulty and are more likely driven by retrieval limitations.

\subsubsection{Manual Validation of Procedural Similarity Judgments}

To assess the reliability of LLM-based relevance scoring, we conducted a manual annotation study on 200 query–trajectory pairs retrieved for unseen tasks. Human annotators labeled each pair as procedurally relevant or irrelevant using the same rubric given to the LLM judge.

We evaluated agreement using standard measures. Humans marked 152/200 trajectories as irrelevant, supporting computation of LLM specificity. Disagreements were also analyzed using Cohen's $\kappa$ to quantify construct divergence between human and model judgments. This study establishes whether LLM scoring offers a consistent and conservative signal for large-scale evaluation.

\FloatBarrier

\section{Experimental Results}

This section presents three complementary evaluation regimes: (1) an exploratory dual-condition analysis on 78 ALFWorld trajectories, (2) a coverage-balanced benchmark on the full 336-trajectory AgentInstruct corpus, and (3) ablation studies examining corpus scale and representation format. These regimes address different questions and should not be compared in absolute MAP values.

\subsection{Exploratory Analysis: Cross-Context Performance Divergence}
\label{sec:exploratory}

We begin with an exploratory evaluation designed to uncover broad method-level behavior under distribution shift. This setup uses 78 trajectories extracted from the ALFWorld environment and a 36-query validation set (18 seen, 18 unseen), without any coverage filtering. All retrieval methods, including lexical baselines, are evaluated here.

\begin{table*}[!htbp]
\centering
\caption{Exploratory dual-condition evaluation (ALFWorld, 78 trajectories). Summary embeddings loses only 11 points while lexical methods degrade sharply ($-$42\%).}
\label{tab:unified_performance}
\begin{tabular}{lcccccc}
\toprule
\textbf{Method} & \multicolumn{2}{c}{\textbf{Seen Tasks}} & \multicolumn{2}{c}{\textbf{Unseen Tasks}} & \textbf{Drop} & \textbf{Rank} \\
\cmidrule(lr){2-3} \cmidrule(lr){4-5}
 & MAP & P@1 & MAP & P@1 & (\%) & Reversal \\
\midrule
Combined Embeddings & \textbf{0.844} & 0.87 & 0.592 & 0.61 & -29.9\% & 1→3 \\
BM25 & 0.815 & 0.83 & 0.469 & 0.45 & \textbf{-42.5\%} & 2→6 \\
Enriched Embeddings & 0.794 & 0.81 & 0.565 & 0.58 & -28.9\% & 3→4 \\
keyword & 0.780 & 0.77 & 0.456 & 0.44 & -41.5\% & 4→5 \\
Action-only Embeddings & 0.756 & 0.74 & 0.488 & 0.49 & -35.5\% & 5→4 \\
Summary Embeddings & 0.754 & 0.72 & \textbf{0.671} & 0.69 & \textbf{-11.0\%} & 6→\textbf{1} \\
\bottomrule
\end{tabular}
\end{table*}
\FloatBarrier
These exploratory results (Table\ref{tab:unified_performance}) expose a clear pattern: methods that rely on lexical features perform well on familiar tasks but degrade sharply in unseen contexts, while object-agnostic abstraction (e.g., \textit{summary embeddings}) achieves comparatively stable performance.

\vspace{0.3em}
\noindent\textbf{Transition.}
While the exploratory analysis uncovers broad generalization failures, it does not control for corpus coverage. To rigorously validate these findings, we next introduce a coverage-balanced benchmark that ensures adequate procedural pattern support for each query. This benchmark requires LLM-based relevance judgments, a computationally expensive procedure that limits evaluation to representative embedding variants.

\subsection{Coverage-Balanced Validation (AgentInstruct, 336 Trajectories)}
\label{sec:coverage_balanced}

To validate these distribution-shift failures under controlled coverage conditions, we evaluate retrieval performance on the full 336-trajectory AgentInstruct corpus using the 40-query coverage-balanced benchmark introduced in §3.4. Only two representative embedding formats are evaluated due to the cost of LLM-as-judge scoring: \textit{state-aware} and \textit{action-only} embeddings.

\textbf{BM25 Exclusion.}  
BM25 is excluded from this benchmark because the coverage-filtering algorithm (Algorithm~\ref{alg:coverage_estimation}) uses keyword matching to require a minimum number of relevant trajectories per query. This introduces information leakage for lexical methods, making BM25 results incomparable under this protocol.

\begin{table}[t]
\centering
\caption{Coverage-balanced benchmark (40 unseen queries, 336 trajectories). Only representative embeddings evaluated due to LLM-as-judge cost.}
\label{tab:coverage_balanced}

\begin{tabular}{lcccc}
\toprule
\textbf{Representation} & \textbf{MAP} & \textbf{EASY} & \textbf{MEDIUM} & \textbf{HARD} \\
\midrule
State-aware & 0.7945 & 0.842 & 0.746 & 0.791 \\
Action-only & 0.7231 & 0.668 & 0.802 & 0.699 \\
\bottomrule
\end{tabular}
\end{table}
\FloatBarrier

\textbf{Interpretation.}  
State-aware embeddings outperform action-only embeddings overall, particularly on simple tasks, but action-only embeddings display competitive or superior performance on medium-complexity procedures, implying that additional contextual text may introduce noise for multi-step patterns.

\subsection{Ablation Studies}
\label{sec:ablation}

\subsubsection{Corpus Size Effects (AgentInstruct: 78 vs.\ 336 Trajectories)}
We evaluate the effect of corpus scale by sampling three 78-trajectory subsets from the AgentInstruct corpus (different random seeds). These subsample evaluations use a separate 20-query set (10 seen, 10 unseen per seed) and are not directly comparable to coverage-balanced results.

\begin{table}[!htbp]
\centering
\caption{Corpus size ablation (AgentInstruct subsamples vs.\ full corpus). Subsample MAP is averaged over three random seeds.}
\label{tab:corpus_ablation}
\begin{tabular}{lccc}
\toprule
\textbf{Corpus} & \textbf{Size} & \textbf{Overall MAP} & \textbf{Retention} \\
\midrule
Full & 336 & 0.7945 & -- \\
Subsample (avg) & 78 & 0.644 & 81.0\% \\
\bottomrule
\end{tabular}
\end{table}
\FloatBarrier
Larger corpus size yields notable improvements in retrieval quality, underscoring the importance of procedural pattern coverage.

\subsubsection{Representation Format Analysis}
Using the coverage-balanced benchmark, we compare state-aware and action-only representations (§4.2). State-aware embeddings benefit simple tasks through additional contextual grounding, while action-only embeddings display robustness in multi-step operations where excessive state descriptions may dilute signal.

\subsection{Validation Studies}
\label{sec:validation}

\subsubsection{Lexical Overlap Analysis}

\textbf{Results:} Relevant query-trajectory pairs display extremely low lexical overlap (2.2--2.8\% shared tokens). Nearly half of high-relevance matches (45\%) occur with Jaccard similarity below 0.05, and correlations between overlap and relevance stay weak ($r=0.29$--$0.40$). These results confirm that relevance judgments arise from procedural similarity rather than lexical matching.
\subsubsection{Semantic Task Space Analysis}

\begin{table}[h]
\centering
\caption{Semantic similarity statistics between task groups.}
\label{tab:semantic_stats}
\begin{tabular}{lcc}
\toprule
\textbf{Comparison} & \textbf{Mean Similarity} & \textbf{Cohen's $d$} \\
\midrule
Seen--Seen & 0.3227 & -- \\
Unseen--Unseen & 0.3227 & -- \\
Seen--Unseen & 0.3186 & 0.026 \\
\bottomrule
\end{tabular}
\end{table}

\begin{figure}[t]
\centering
\includegraphics[width=\columnwidth]{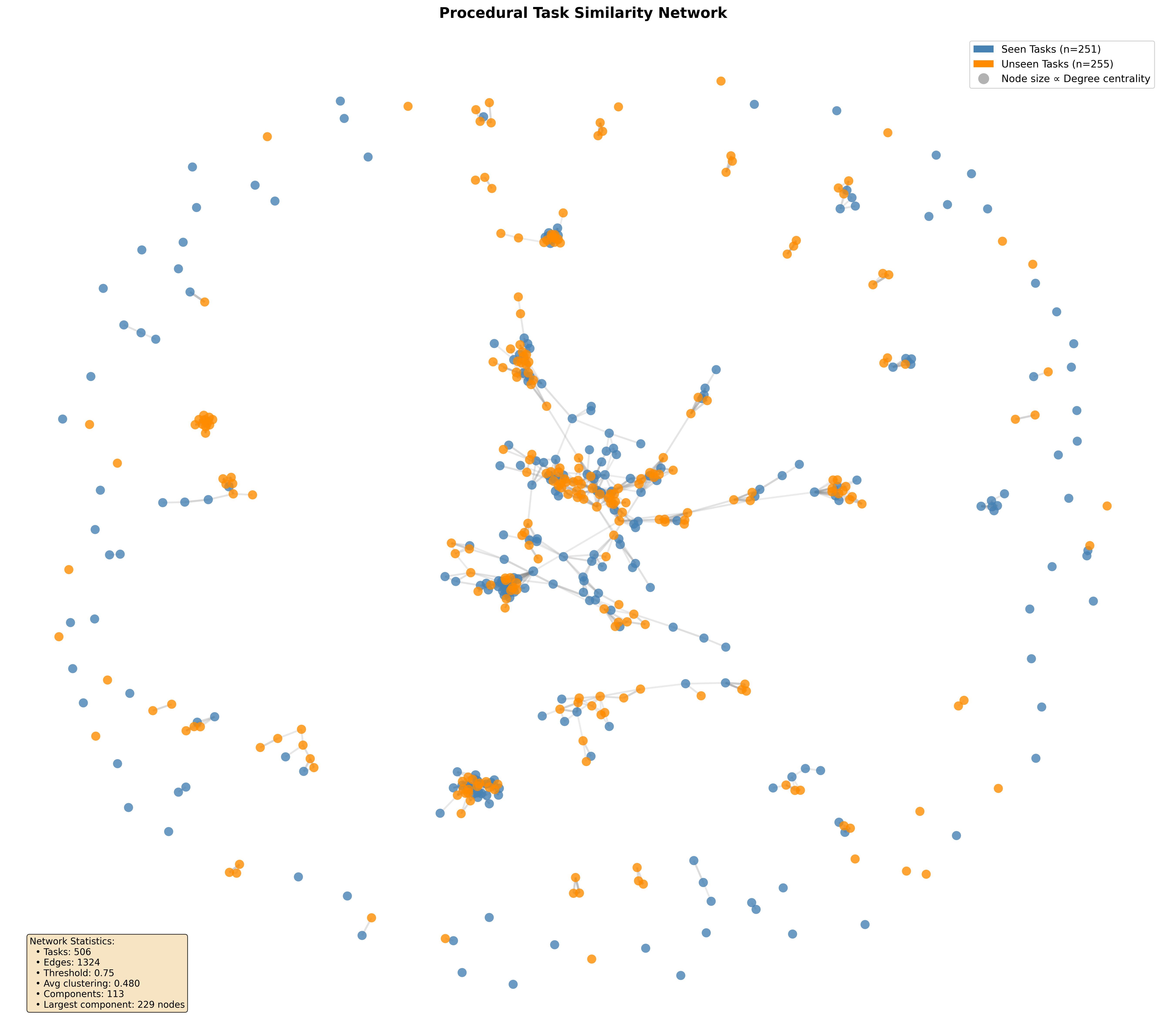}
\caption{t-SNE visualization of 327 ALFWorld tasks. Seen and unseen tasks display weak separation (Cohen's $d\approx 0.02$-$0.04$), pointing to considerable semantic overlap.}
\label{fig:tsne}
\end{figure}

\textbf{Results:} The semantic structure of ALFWorld tasks displays minimal separation between seen and unseen groups as documented in Figure~\ref{fig:tsne}. Within-group similarities (0.3227) nearly match cross-group similarity (0.3186), yielding a negligible separation of 0.004 (1.27\%) and extremely small effect size (Cohen's $d=0.026$). The Silhouette score is near zero ($s=0.0059$), confirming no meaningful clustering. These metrics establish that both groups occupy a shared, homogeneous semantic space.

\subsubsection{Manual Validation Study}

We manually annotated 200 query–trajectory pairs retrieved for unseen queries to test whether LLM-based relevance scoring captures human judgments.

\begin{figure}[t]
\centering
\includegraphics[width=0.7\columnwidth]{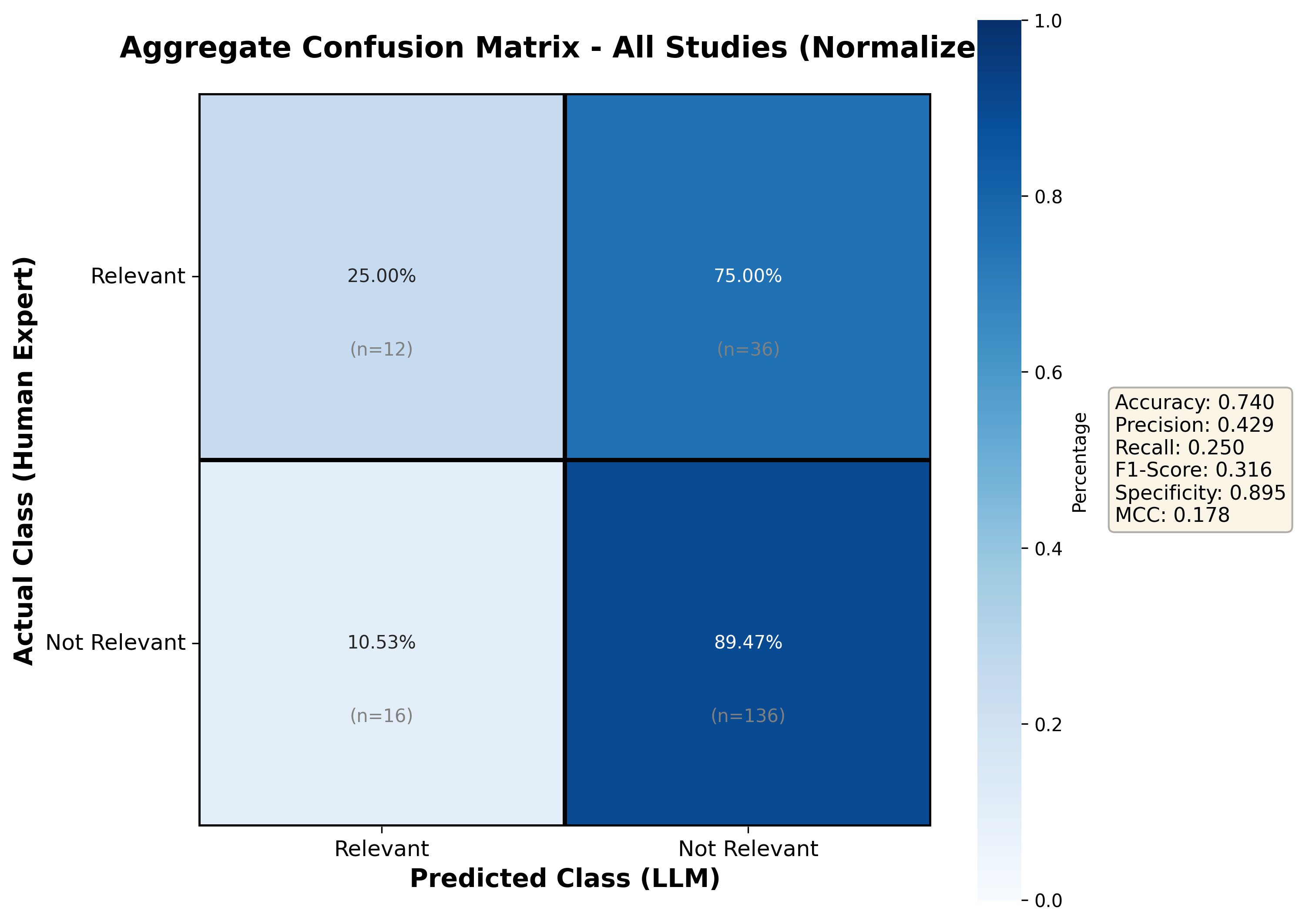}
\caption{Human-LLM agreement on procedural relevance judgments (binary threshold at 6/10). High specificity confirms reliable failure detection, while disagreements arise primarily from partial procedural utility recognized by humans.}
\label{fig:manual_validation}
\end{figure}

Humans labeled 152/200 pairs as irrelevant (Figure-\ref{fig:manual_validation}). The LLM achieved 89.5\% specificity on these cases, confirming that it reliably identifies retrieval failures. Divergence appeared primarily on borderline cases, with moderate construct mismatch (Cohen's $\kappa = 0.178$): humans often recognized partial procedural utility that the LLM scores below threshold, not the reverse. We use the judge only to compute relative comparisons spanning methods, where its bias is stable and affects all systems uniformly. This asymmetry means the LLM judge applies stricter structural criteria, whereas humans accept adaptable sub-procedures. These results establish that LLM-based scoring offers a conservative, reproducible lower bound on relevance while preserving the core signal: embedding methods retrieve mostly irrelevant trajectories when object vocabularies shift, reinforcing the generalization cliff observed throughout experiments.

\section{Discussion}

\subsection{Architectural Origins of Distribution-Shift Failures}

Throughout all experimental regimes, we observe a consistent 25-40\% degradation when systems encounter procedurally novel contexts. These systematic failures manifest as both a reversal in method rankings and considerable MAP drops. Its consistency spanning corpora, evaluation setups, and representation formats means it arises from architectural constraints rather than dataset artifacts.

A key mechanism emerges from our controlled ablations: sentence-transformer encoders treat trajectories as unordered collections of tokens due to mean pooling, which averages token embeddings and discards temporal structure. Whether representing compact action sequences or state-rich descriptions, the encoder produces embeddings that are nearly indistinguishable for procedurally different trajectories containing similar vocabularies.
Although contextual encoders incorporate positional information, mean pooling attenuates much of this signal in practice, reducing sensitivity to temporal ordering.

This limitation can be illustrated formally. Consider two sequences with identical tokens but different orderings:
\begin{align*}
S_1 &= [a_1 \rightarrow a_2 \rightarrow a_3], \\
S_2 &= [a_2 \rightarrow a_1 \rightarrow a_3].
\end{align*}
Under mean pooling with token embeddings $e(w_i)$:
\begin{equation}
\text{embed}(S) = \frac{1}{n}\sum_{i=1}^{n} e(w_i),
\end{equation}
the representations of $S_1$ and $S_2$ differ only by permutation of terms inside the sum, making them nearly identical in practice. The encoder cannot reliably distinguish valid multi-step procedures (e.g., ``take apple, then heat'') from invalid ones with inverted ordering. This architectural bottleneck explains why medium-complexity tasks which rely most heavily on temporal dependencies face the largest generalization gaps.

It also clarifies why adding 10$\times$ more contextual information in state-aware embeddings results in modest gains. Because mean pooling dilutes the contribution of individual actions, additional state tokens shift the aggregate representation but do not give the model temporal reasoning capabilities. In our benchmark, this leads to behavior such as action-only embeddings outperforming state-aware representations on medium-complexity tasks, despite using far fewer tokens.

\subsection{Why LLM-Generated Abstractions Generalize Better}

Procedural summaries generated by LLMs achieve considerably higher generalization resilience not because the downstream embedding model is better, but because the LLM produces abstracted representations that decouple procedures from specific object identities. Summaries like ``locate item $\rightarrow$ heat item $\rightarrow$ place in storage'' stress structural patterns rather than surface-form differences. This preprocessing step lets the sentence transformer operate on higher-level procedural concepts, partially compensating for its inability to encode temporal relations directly.

Thus, the strength of \textbf{Summary Embeddings} stems from its generation process rather than its embedding mechanism: the LLM explicitly extracts procedural structure before similarity computation.

\subsection{The Memorization-Generalization Trade-off}

The complete rank inversion between seen and unseen conditions captures a fundamental trade-off in current architectures. Methods optimized for memorization excel on seen tasks (79-86\% MAP) but retain only 55-70\% of this performance when object vocabularies shift. Yet they suffer 30-42\% drops on unseen tasks where such cues are unavailable. Abstraction-oriented methods maintain stability spanning contexts but sacrifice some in-distribution performance.

This trade-off appears to arise from single-stage retrieval pipelines that conflate representational abstraction with similarity computation. Without explicit modeling of structure, increasing representational richness cannot simultaneously support both memorization and broad generalization.

\subsection{Implications for Retrieval System Design}

Our results point to several practical guidelines for designing procedural retrieval systems:

\textbf{Corpus scale has larger impact than representation enrichment.} A 4.3$\times$ increase in corpus size yields a 27.7\% improvement, compared to a 9.9\% gain from adding richer state context. Enhancing procedural coverage is more influential than architectural tweaks within sentence-transformer-based pipelines.

\textbf{Returns diminish with corpus growth.} Gains plateau as the corpus expands beyond 100 trajectories for common patterns, consistent with diminishing-returns behavior where additional examples increasingly repeat existing structures.

\textbf{Match retrieval strategies to deployment needs.} Memorization-oriented models perform best in stable domains with repeated object vocabularies, while abstraction-oriented approaches suit environments with frequent novelty.

\textbf{Structured architectures look promising.} A promising direction is a two-stage design separating (1) procedural structure extraction from (2) similarity computation. This avoids forcing static embeddings to encode both content and order simultaneously, a known limitation of mean-pooled encoders.

\subsection{Limitations}

Several limitations should be noted. LLM-as-judge evaluation incurs computational cost, restricting coverage-balanced experiments on the 336-trajectory corpus to representative variants. Human-LLM agreement is moderate, with low $\kappa$ meaning human annotators identify partial procedural utility that conservative LLM scoring may overlook; our results thus represent lower bounds on practical utility. Our benchmark focuses on ALFWorld's household tasks, and while the underlying properties of procedural structure are domain-general, verification in robotics, software workflows, or multi-agent settings is still needed. Finally, our study evaluates retrieval at inference time without fine-tuning or architectural adaptation, which may influence representational behavior.

\subsection{Future Directions}

Several avenues emerge from our findings. Architecturally, encoders that preserve temporal dependencies such as sequence-aware transformers, hierarchical trajectory encoders, or causal-attention mechanisms could better represent procedural structure than mean-pooled models. Representing trajectories as temporal knowledge graphs offers another promising direction, capturing action dependencies and state transitions explicitly. Compositional retrieval, where systems assemble procedures from reusable components, may support stronger generalization to novel tasks. Dynamic corpus expansion through online learning could uncover new scaling behaviors as agents accumulate experience. Finally, integrating human feedback with retrieval pipelines may help systems learn to recognize partial procedural relevance beyond what current encoders can detect.

\section{Conclusion}
\label{sec:conclusion}

We introduced the first systematic benchmark for evaluating procedural memory retrieval, isolating an agent's ability to recognize structurally similar trajectories spanning differing object contexts. Using six retrieval methods evaluated on two complementary corpora (78 expert demonstrations and 336 LLM-generated trajectories) and 40 coverage-balanced queries, we uncovered the generalization cliff: methods that achieve strong in-distribution performance (84\% MAP) suffer marked degradation on novel contexts (59\% MAP, $-$30\% drop), while LLM-derived procedural abstractions \textbf{degrade minimally, losing only 11 percentage points}

Throughout five validation studies, we established that sentence-transformer encoders treat procedural text as unordered token sets, even when enriched with state information. As a result, adding 10× more contextual detail yields only modest gains (9.9\%), and action-only representations outperform state-aware variants on medium-complexity tasks. These findings suggest that architectural constraints—rather than corpus scale or representation richness—set the ceiling for current embedding-based retrieval.

Our benchmark differentiates genuine procedural understanding from vocabulary-driven pattern matching, supporting clearer diagnosis of memorization versus abstraction behaviors. The 3× difference in generalization resilience spanning methods offers practical guidance: memorization-oriented systems suit stable environments, whereas abstraction-oriented approaches are needed when tasks or object vocabularies shift.

We release all corpora, the coverage-balanced query bank, and evaluation protocols to support progress toward retrieval systems that generalize procedural knowledge reliably spanning contexts.

\bibliographystyle{unsrtnat}
\bibliography{references}

\end{document}